\documentclass[sigconf,natbib=false]{acmart}

\AtBeginDocument{%
  }

\setcopyright{acmlicensed}
\copyrightyear{2024}
\acmYear{2024}
\acmDOI{XXXXXXX.XXXXXXX}

\acmConference[KDD-UC'24]{}{August 25--29,
    2024}{Barcelona, Spain}

\acmISBN{978-1-4503-XXXX-X/18/06}

\RequirePackage[
  datamodel=acmdatamodel,
  style=acmnumeric,
  ]{biblatex}

\begin{document}

\title{Towards Measuring Sell Side Outcomes in Buy Side Marketplace Experiments using In-Experiment Bipartite Graph}

\author{Vaiva Pilkauskaitė}\authornote{Leading author, work completed in partial fulfillment of the requirements for a Bachelor of Mathematics degree at Vilnius University.}
\email{vaiva.pilkauskaite@gmail.com}
\affiliation{%
  \institution{Vilnius University}
  \city{Vilnius}
  \country{Lithuania}
}

\author{Jevgenij Gamper}
\email{jevgenij.gamper@vinted.com}
\affiliation{%
  \institution{Vinted}
  \city{Vilnius}
  \country{Lithuania}
}

\author{Rasa Giniūnaitė}
\email{rasa.giniunaite@vinted.com}
\affiliation{%
  \institution{Vilnius University, Vinted}
  \city{Vilnius}
  \country{Lithuania}
}

\author{Agnė Reklaitė}
\email{agne.reklaite@vinted.com}
\affiliation{%
  \institution{Vilnius University, Vinted}
  \city{Vilnius}
  \country{Lithuania}
}

\renewcommand{\shortauthors}{Pilkauskaitė et al.}

\begin{abstract}

In this study, we evaluate causal inference estimators for online controlled bipartite graph experiments in a real marketplace setting. Our novel contribution is constructing a bipartite graph using in-experiment data, rather than relying on prior knowledge or historical data, the common approach in the literature published to date. We build the bipartite graph from various interactions between buyers and sellers in the marketplace, establishing a novel research direction at the intersection of bipartite experiments and mediation analysis. This approach is crucial for modern marketplaces aiming to evaluate seller-side causal effects in buyer-side experiments, or vice versa. We demonstrate our method using historical buyer-side experiments conducted at Vinted, the largest second-hand marketplace in Europe with over 80M users.
\end{abstract}


\received{27 May 2024}
\received[accepted]{19 June 2024}

\maketitle

\section{Introduction}
Technology companies have adopted experimentation, as the primary method for software and product development. For instance, Microsoft's Bing team conducts over 10,000 experiments annually \cite{kohavi2020trustworthy, fabijan2017evolution, fabijan2017benefits}. This large-scale adoption necessitates the implementation of automated experiment analysis platforms that integrate scientific rigour with engineering efficiency to estimate treatment effects accurately, such as changes to user interface elements or recommendation algorithms \cite{gupta2018anatomy, xu2015infrastructure, diamantopoulos2020engineering}. 

In social networks and marketplaces experiment designs extend beyond simple randomised control trials \cite{xu2015infrastructure, eckles2017design, saveski2017detecting, blake2014marketplace, fradkin2019simulation, holtz2020reducing}, where the response of any
experiment unit (user) under treatment is dependent on the response of another unit under treatment. This is known as violation of the Stable Unit Treatment Value Assumption (SUTVA) or interference \cite{hernan2020causal}. In two sided marketplaces the level of interference depends on the supply and demand dynamics \cite{johari2022experimental, li2021interference}. 

An engineering team in a marketplace setting, may want to run a buy side experiment, yet want to know the causal effect on the sell side. For instance, they might test a feature to increase buyer engagement and wish to quantify whether this intervention positively impacts seller outcomes without causing adverse affects. This scenario fits well within a bipartite experiment framework, where we assign treatments to one set of units (buyers) and measure outcomes for another set of units (sellers). The two sets of units are connected by a bipartite graph, governing how the treated units can affect the outcome units. 

Bipartite experiment analysis methods published to date, have only considered graphs based on pre-experimental data. In contrast, in this work we use graphs constructed using in-experiment data using various interaction events. We test various inference methods proposed to date, and propose new inference variants. We also show that bipartite graphs constructed using different interaction events lead to different estimates - signaling a promising research direction at the intersection between bipartite experiments and mediation analysis.

This work is organized as follows. In Section \ref{prelim:basics} we introduce randomized experiments, potential outcome framework as well key notation and terms for bipartite experiments. We discuss related work in Section 
\ref{ref:past-work}. In Section \ref{ref:bipartite-graph} we build up intuition on how bipartite graphs can be constructed using in-experiment data in a large marketplace. We then introduce the considered inference methods and review their results in Sections \ref{ref:methods} and \ref{ref:experimental_data_results}, and discuss future work in Section \ref{ref:conclusions}.

\section{Preliminaries}
\label{prelim:basics}

In experiments conducted by technology companies units (e.g., users) are randomly assigned to treatment or control groups, and their outcomes are averaged and compared between the groups, ensuring unbiased estimates of treatment effects by mitigating confounding factors. This method is widely used to infer the causal impact of various changes in user interfaces, recommendation algorithms, and other features to enhance user engagement and satisfaction \cite{kohavi2020trustworthy}.

The potential outcomes framework is fundamental for causal inference in experiment designs, including bipartite experiments. For each unit \(i\), we define \(Y_i(Z=1)\) as the outcome if the unit receives the treatment (where \(Z=1\) indicates treatment) and \(Y_i(Z=0)\) as the outcome if the unit does not receive the treatment (where \(Z=0\) indicates control). The causal effect for unit \(i\) is then \(Y_i(1) - Y_i(0)\). Since we can only observe one of these outcomes for each unit, the challenge lies in estimating the unobserved potential outcome. By leveraging randomization, we can estimate the average treatment effect (ATE), denoted as \(\tau\), which is the difference in mean outcomes between the treatment and control groups. Formally, the ATE is defined as \(\tau = E[Y_i(1)] - E[Y_i(0)]\), where \(E\) denotes the expectation \cite{hernan2020causal}.

\begin{figure}
  \centering
  \includegraphics[width=\linewidth]{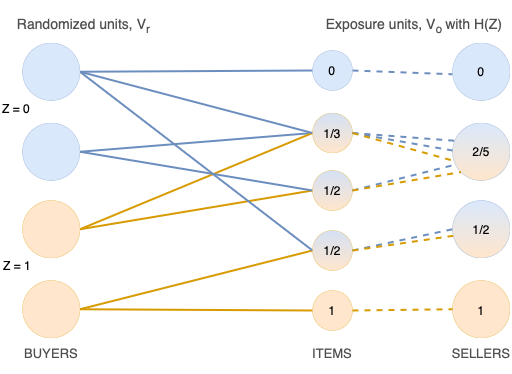}
  \caption{Bipartite Graph adapted for Vinted Marketplace, treated buyers can interact with many sellers and their items, thus violating the SUTVA assumption. Exposure Score $H_i(Z)$ (\ref{eq:exposureH}) can be calculated for both Items and Sellers.}
  \Description{Bipartite Graph adapted for Marketplace Settings}
  \label{fig:bipartite}
\end{figure}

Bipartite experiment, further builds on the above setup by having two distinct groups of units: the \textit{diversion} units, which are randomized to receive the treatment, and the \textit{outcome} units, where the outcomes are measured. Unlike traditional experiments where these groups are the same, bipartite experiments separate them. This dependence is represented by a bipartite graph linking diversion units to outcome units. In the bipartite graph, the set of \(m\) diversion units is denoted as \(V_r\) and the set of \(n\) outcome units as \(V_o\). Each diversion unit receives a random binary treatment \(Z_i\) (0 or 1), forming a vector \(Z = (Z_1, Z_2, \ldots, Z_m)\) where \(Z \in \{0, 1\}^m\). Each outcome unit \(i\) in \(V_o\) has a potential outcome function \(Y_i(Z)\), which maps the treatment assignments to the observed value. 

Causal inference of ATE in a bipartite experiment relies on two assumptions \cite{harshaw2023design}. First, \textit{Linear Exposure Assumption} states that the treatment assignments influences the potential outcomes only through a linear combination. More formally, for each outcome unit $i \in V_o$, the exposure $H$ of outcome unit $i$ is
\begin{equation}
\label{eq:exposureH}
   H_i(Z) = \sum_{r=1}^m w_{i,r}Z_r 
\end{equation}

Where, $\forall i$, $\sum w_{ir} = 1$, i.e. the weights incident to an outcome unit are normalized to sum to one. 

Second, \textit{Linear Response Assumption} suggests that the potential outcome for each outcome unit is a linear function of its exposure. This means for each outcome unit $i$ in $V_o$, there are specific parameters $\alpha_i$ (intercept) and $\beta_i$ (slope) such that
\begin{equation}
\label{eq:linear_exposureY}
     Y_i(Z) = \alpha_i + \beta_i H_i(Z) 
\end{equation}
The values of $\alpha_i$ and $\beta_i$ vary between units and are unknown to the experimenter. The experimenter only sees the outcome $Y_i(Z)$, the treatment assignment vector $Z$, and the exposure $H_i$.

Under the linear exposure-response model, the ATE can be expressed as follows:
\begin{equation}
\tau = \frac{1}{n} \sum_{i=1}^n \beta_i.
\label{eq:atte_beta}
\end{equation}

\section{Related Work}
\label{ref:past-work}
Using bipartite graphs to estimate causal impact of treatment to one side of units to another was formally introduced by Zigler et al. \cite{zigler2018bipartite}, to measure the impact of power plants' pollution prevention treatments on hospitalization rates at surrounding hospitals.

Harshaw et al. \cite{harshaw2023design} developed bipartite experiments theory further specifically for two-sided marketplace settings to mitigate SUTVA violations. Notably, proposing the first reliable inference method in bipartite settings - the Exposure-Reweighted Linear (ERL) and variance estimators that are consistent, unbiased and asymptotically normal in sufficiently sparse bipartite graphs. Shi et al. \cite{shi2024scalable} further extend previous work by proposing a covariate-adjusted ERL (CR-ERL) estimator to reduce variance, and develop scalable inference methods to compute the estimate. 

In this work we use both ERL and CR-ERL estimators. However, in all of the literature published to date, the structure of a bipartite graph was fixed before the experiment. Yet our bipartite graph is not known before the experiment. 

\section{Constructing bipartite graph using in-experiment data}\label{ref:bipartite-graph}
Let us now motivate how a bipartite experiment may be constructed using in-experiment data in a real marketplace setting such as Vinted. Vinted is a largest in Europe marketplace for second-hand items, where a user can upload items to sell, as well as purchase items from other users. 

Product engineering teams at Vinted routinely conduct experiments on buy or sell side of the marketplace, and would benefit from bipartite experiment design. For the rest of this work we will focus on a scenario where a team is conducting a buyer-side experiment, and would like to measure seller-side outcomes. To construct a bipartite graph in this setting we set the diversion units as buyers who receive the treatment, and sellers as outcome units. 

Unlike in the literature published to date, one cannot know in advance which sellers the randomized buyers will engage with. Therefore to establish links between the diversion and outcome units note that users in Vinted marketplace can interact with each other in different ways. For example, buyers can view sellers' listings, favorite them, send messages or offers, purchase items, or visit seller profiles. We use item views and favorite events in order to construct the links. Typically, buyers interact with sellers by engaging with the sellers’ items, where the latter can also be set as outcome units units in the bipartite graph, however in this work we focus specifically on sellers as outcome units (see Figure \ref{fig:bipartite}). 

\begin{figure}
  \centering
  \includegraphics[width=\linewidth]{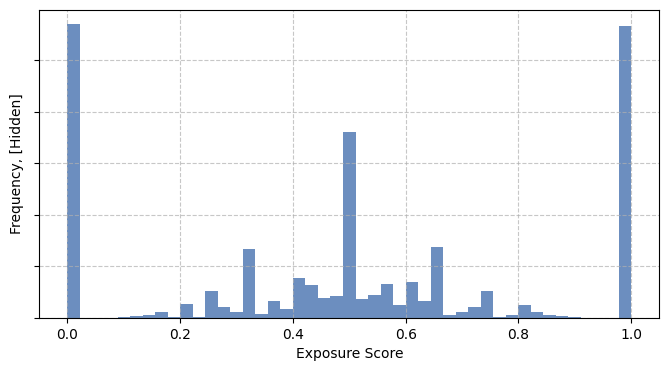}
  \caption{Sellers' Exposure Score distribution example for \textit{On} and \textit{Off} experiment. The distribution depends on how many interactions between buyers and sellers happened, as well as on the probability of buyers being from different test variants. With many sellers items receiving, for instance, one interaction from buyer, fewer receiving two, and so on - it is expected that an exposure frequency of 0 or 1 is the most common, followed by 0.5}
  \Description{Exposure Score distribution, histogram}
  \label{fig:distrib}
\end{figure}

After running the experiment and mapping the bipartite graph using item views or favorites, we calculate the exposure $H_i(Z)$ for each outcome unit (seller) based on the linear exposure assumption (\ref{eq:exposureH}). If there are only two test variants (i.e., \textit{On} and \textit{Off}), we set \(Z = 0\) for buyers in the \textit{Off} variant and \(Z = 1\) for those in the \textit{On} variant. We present an example of exposure distribution constructed using in-experiment data in Figure \ref{fig:distrib}). Notice that many sellers have only a single interaction, resulting in an exposure score of either 0 or 1. For these sellers, the SUTVA assumption holds true. Similarly, sellers who are exposed to a single group of buyers, also get an exposure  of either 0 or 1. 

In case of multiple variants in the same experiment, where we need to evaluate \textit{Off} vs \textit{A}, and \textit{Off} vs \textit{B} we would construct a separate graph for each variant. Hence, we would have a vector of exposures for each specific variants ($H_i^A, H_i^B, ... , H_i^N$). The second method for tackling multiple variants involves using a normalized Exposure Score to account for interactions from buyers across all variants in the experiment. For instance, if a seller received one interaction from the Off group, two from the A group, and three from the B group, the exposure scores would be as follows: $H_i^A = 1/3$, $H_i^B = 1/2$. This approach ensures that the influence of each variant is proportionally represented in the exposure score and this can be advantageous when the number of interactions varies significantly between variants, as it mitigates the risk of skewing the results due to disproportionate interaction counts.

\section{Effect Estimation Methods}\label{ref:methods}
In this section, we leverage the three different approaches to estimate causal impact of an intervention in a bipartite experiment setting where the buyer side receives a treatment, and the outcomes are measured on the seller side. We also elaborate on the difficulties encountered with real, large-scale industry experiments.

\subsection{First Approach: The Exposure-Reweighted Linear Estimator}
One of the first methods we test on real experimental is Exposure-Reweighted Linear (ERL) introduced by Harshaw et al. \cite{harshaw2023design}. The estimator is expressed as follows:
\begin{equation}\label{eq:ERL}
    \hat{\tau} = \frac{1}{n} \sum_{i=1}^n Y_i(Z) \left(\frac{H_i(Z) - E[H_i(Z)]}{\operatorname{Var}[H_i(Z)]}\right)
\end{equation}

$\hat{\tau}$ here is an estimate for Average Treatment Effect (ATE). Where the exposure $H_i(Z)$ and metrics $Y_i(Z)$ were calculated for every seller in the bipartite graph constructed using either item view or favorite events that happened during the experiment. 

To estimate variance of the ATE, Harshaw et al. \cite{harshaw2023design} proposed the following estimator of the ERL variance:.

\begin{equation} \label{eq:ERL Var}
  \widehat{\operatorname{Var}}(\hat{\tau}) = \frac{1}{n^2} \sum_{i=1}^{n} \sum_{j=1}^{n} \hat{C}_{i,j} = \frac{1}{n^2} \sum_{i=1}^{n} \sum_{j=1}^{n} Y_i Y_j R_{i,j}(H_i, H_j) 
\end{equation}
where the weighting estimators $\hat{C}_{i,j}$ are used to evaluate $\operatorname{Cov}(\hat{\tau_i}, \hat{\tau_j})$ terms where
$Y_iY_j$ is the product of observed outcomes and $R_{i,j}(H_i, H_j)$ is a weighting function which takes the exposures as inputs. It is stated that under the linear response assumption, and assumption that for each pair of outcome units $i, j \in [n]$, the covariance matrix $\Sigma_{i,j}$ of their exposures $H_i$ and $H_j$ satisfies the non-degeneracy condition, ensuring $\det(\Sigma_{i,j}) > 0$, the variance estimator of the ERL point estimator is unbiased.

Despite the proposed variance estimator, we opted in for a simple bootstrapping based approach for estimating the confidence intervals for ERL estimator \cite{stine1989introduction}. Reason being that proposed variance estimator (\ref{eq:ERL Var}) computational complexity is $\Omega(n^2)$ because it involves a double summation over n units as it calculates pairwise interactions between all exposures which is not suitable for the amount of data in Vinted's marketplace - tenths of millions in an experiment. Second, if we were to even consider smaller samples of data, variance estimator (\ref{eq:ERL Var}) is unbiased only when non-degeneracy conditions hold as described in the preliminaries. In our bipartite graphs we found that the necessary conditions do not hold as two outcome units can have identically weighted edges. While the Harshaw et al. \cite{harshaw2023design} proposes an alternative variance estimator that adjusts the weighting function to handle cases where $\det(\Sigma_{i,j}) = 0$, computational complexity issue remains. 

\subsection{Second Approach: Regression Based Estimator}
As stated in Section \ref{prelim:basics} under the linear response assumption (\ref{eq:linear_exposureY}), regression-based estimand is equal to the average total treatment effect. Applying linear regression to estimate the ATE is compelling due to its simplicity and flexibility, such as using pre-experimental outcomes in order to decrease ATE variance \cite{deng2013improving}. 

Our regression is based on the following linear relationship: 
\begin{equation}
\label{eq:linear_exposureY}
     Y^{in}_i = \alpha + \tau \cdot H_i(Z) + \psi \cdot Y^{pre}_{i}
\end{equation}
Where $\alpha$ is constant, and $\tau$ is the ATE estimate that we report, while $Y^{in}_i$, $Y^{pre}_{i}$, $H_i(Z)$ are seller measured outcomes during the period of an experiment, seller outcomes before the experiment begins and their exposure, respectively. For variance reduction, we incorporated the pre-experimental period seller measured outcomes \cite{deng2013improving}.

\subsection{Third Approach: Covariate-adjusted Exposure Reweighted Linear Estimator}
The final approach we tested on real world experiment data was the Covariate-adjusted Exposure Reweighted Linear (CR-ERL) estimator proposed by Shi et al. \cite{shi2024scalable}. CR-ERL extends linear exposure-response model by incorporating the pre-experimental outcomes into the estimator. The advantage of this method was that the proposed inference method, based on Monte Carlo randomization, is easy to implement. The estimator takes the following form:
\begin{equation} \label{eq:ca-erl}
    \hat{\tau}(\lambda) = \frac{1}{n} \sum_{i=1}^n (Y_i(Z) - \lambda f_i(X_i))\left(\frac{H_i(Z) - E[H_i(Z)]}{\operatorname{Var}[H_i(Z)]}\right)
\end{equation}
where $f_i(X_i)$ is a function of the covariates, and $\lambda$ is a parameter that controls for the estimator’s variance. 

The resulting confidence intervals were the narrowest among all methods we tested.

\begin{figure}
  \centering
  \includegraphics[width=\linewidth]{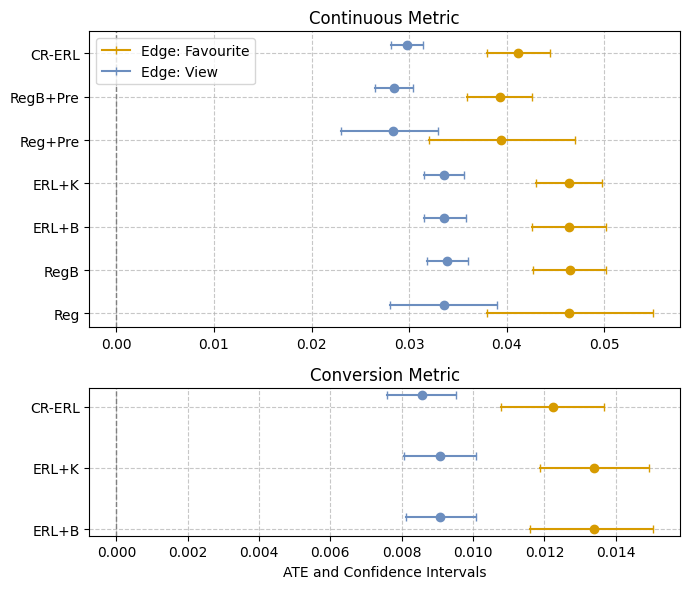}
  \caption{Comparison of estimates and confidence intervals for different methods applied to two bipartite graphs with different buyer interaction events in the same experiment for two metrics (continuous and conversion). Methods include: Reg (Regression), RegB (Bootstrapped Regression), ERL+B (Bootstrapped ERL Confidence Intervals), ERL+K (Confidence Interval Estimation with Randomization Algorithm). "Pre" indicates the inclusion of a pre-experiment covariate.}
  \Description{results}
  \label{fig:results1}
\end{figure}

\section{Experimental data and results}\label{ref:experimental_data_results} 
To evaluate the above methods in a setting where a bipartite graph is constructed using in experimental data, we took a buyer-side experiment in Vinted marketplace. The challenge remains such that the true treatment effect is unknown, therefore we selected an experiment where there is a detected buyer-side effect, and there is a strong qualitiative evidence for sell-side effects. This experiment sample size consisted of approximately 20 million users, highlighting the importance of computational efficiency as a factor when selecting the best estimator.

Our findings using the above presented estimators and their variants are shown Figure \ref{fig:results1}. The top element of the figure corresponds to treatment effects estimated for continuous sell-side metric, and the bottom corresponds to conversion sell-side metric. We demonstrate two metrics as continuous and conversion metrics have different sensitivity. As expected, conversion metric showed narrower confidence intervals and smaller effect size in general as those metrics are generally less susceptible to outliers and show smaller variability due to binary nature. 

All of the above methods were implemented using two different bipartite graphs using edges based on different types of buyer interactions. First bipartite graph edges are constructed using item views, and the second graph constructs edges based on favorites. Item views are the primary means by which buyers connect with sellers, as they are the initial interaction that enables all subsequent actions: a favorite event can only occur if an item view has already taken place. Notice that the estimates using bipartite graph created with views events estimated significantly lower average treatment effect compared to favorites events for all the methods and both metrics. This raises a question of potential mediation effects from buyers, to their change in viewing or favoring items, to seller side outcomes - where the exposure itself is a mediator. However, we leave the exploration of this for future work.

To begin our study of the above presented estimators, let us focus on the second approach, a regression based estimate and its variants, labeled as Reg (Regression), RegB (Bootstrapped Regression), ERL+B (Bootstrapped ERL Confidence Intervals), ERL+K (Confidence Interval Estimation with Randomization Algorithm). We found that the estimated coefficient closely matched the first approach, an ERL estimate proposed by Harshaw et al. \cite{harshaw2023design}. The question is whether we could trust the confidence intervals, given that the diagnostic regression residual plots indicated heteroskedasticity and non-normality. Although these assumptions appear to be violated, it is worth noting that with a large amount of data, these assumptions are less critical due to the robustness of regression estimates in large samples. To assess the validity of the regression confidence intervals, we also employed bootstrapping to estimate the regression confidence intervals, labelled as RegB. Results from regression showed the highest uncertainty with widest confidence intervals. Although, regression estimated coefficients were consistent with other methods. Bootstrapping ERL performed similarly to the ERL Monte Carlo randomization method, providing fast and comparable results instead of the complex variance estimator suggested by \cite{harshaw2023design}. Adjusting for pre-experiment data showed lower estimates for all methods, which may be due to heavy sellers that create different baselines before the experiment. CR-ERL based estimates overall show the most satisfactory results. By leveraging pre-experiment data, we conclude that we can be more confident that the average treatment effect is not overestimated. As both bootstrapped and randomization-based inference with ERL showed similar results, they can be used to validate consistency between the two methods in specific experiments. These techniques are recommended when pre-experiment metrics are difficult to obtain. While regression is the least computationally expensive method, other approaches are preferable due to stronger theoretical guarantees.

\section{Conclusions and Future Work}\label{ref:conclusions}
In this work we demonstrated the use of in-experiment data such as viewing and favorite events in order to construct a bipartite graph and obtain causal effect estimates using methods proposed in the literature. We compare the precision of different estimators on a real world experimental data obtained from Vinted, a largest secondhand marketplace in Europe. Our results, specifically the difference in the estimated effects between views and favorites based bipartite graphs, raise interesting new research avenues for combining mediation analysis and bipartite experiment inference methods. In summary:
\begin{itemize}
    \item Mediator selection for bipartite graph construction is important as it can have significant effect on the magnitude of estimates.
    \item It appears practically feasible to use in-experiment data to construct exposure scores for causal treatment effect inference in a bipartite experiment design.
    \item Our results demonstrate that non-parameteric methods such as CR-ERL method proposed by Shi et al. \cite{shi2024scalable} should be preferred. 
\end{itemize}
We leave it for future work to perform a meta-analysis of these methods and mediators with a large dataset of historical experiments, so that systemic differences in the estimators performance could be studied \cite{tripuraneni2021meta}.

\printbibliography

@book{kohavi2020trustworthy,
  title={Trustworthy online controlled experiments: A practical guide to a/b testing},
  author={Kohavi, Ron and Tang, Diane and Xu, Ya},
  year={2020},
  publisher={Cambridge University Press}
}

@inproceedings{gupta2018anatomy,
  title={The anatomy of a large-scale experimentation platform},
  author={Gupta, Somit and Ulanova, Lucy and Bhardwaj, Sumit and Dmitriev, Pavel and Raff, Paul and Fabijan, Aleksander},
  booktitle={2018 IEEE International Conference on Software Architecture (ICSA)},
  pages={1--109},
  year={2018},
  organization={IEEE}
}

@inproceedings{xu2015infrastructure,
  title={From infrastructure to culture: A/B testing challenges in large scale social networks},
  author={Xu, Ya and Chen, Nanyu and Fernandez, Addrian and Sinno, Omar and Bhasin, Anmol},
  booktitle={Proceedings of the 21th ACM SIGKDD International Conference on Knowledge Discovery and Data Mining},
  pages={2227--2236},
  year={2015}
}

@inproceedings{diamantopoulos2020engineering,
  title={Engineering for a science-centric experimentation platform},
  author={Diamantopoulos, Nikos and Wong, Jeffrey and Mattos, David Issa and Gerostathopoulos, Ilias and Wardrop, Matthew and Mao, Tobias and McFarland, Colin},
  booktitle={Proceedings of the ACM/IEEE 42nd International Conference on Software Engineering: Software Engineering in Practice},
  pages={191--200},
  year={2020}
}

@inproceedings{fabijan2017evolution,
  title={The evolution of continuous experimentation in software product development: from data to a data-driven organization at scale},
  author={Fabijan, Aleksander and Dmitriev, Pavel and Olsson, Helena Holmstr{\"o}m and Bosch, Jan},
  booktitle={2017 IEEE/ACM 39th International Conference on Software Engineering (ICSE)},
  pages={770--780},
  year={2017},
  organization={IEEE}
}

@inproceedings{fabijan2017benefits,
  title={The benefits of controlled experimentation at scale},
  author={Fabijan, Aleksander and Dmitriev, Pavel and Olsson, Helena Holmstr{\"o}m and Bosch, Jan},
  booktitle={2017 43rd Euromicro Conference on Software Engineering and Advanced Applications (SEAA)},
  pages={18--26},
  year={2017},
  organization={IEEE}
}

@article{eckles2017design,
  title={Design and analysis of experiments in networks: Reducing bias from interference},
  author={Eckles, Dean and Karrer, Brian and Ugander, Johan},
  journal={Journal of Causal Inference},
  volume={5},
  number={1},
  year={2017},
  publisher={De Gruyter}
}

@inproceedings{saveski2017detecting,
  title={Detecting network effects: Randomizing over randomized experiments},
  author={Saveski, Martin and Pouget-Abadie, Jean and Saint-Jacques, Guillaume and Duan, Weitao and Ghosh, Souvik and Xu, Ya and Airoldi, Edoardo M},
  booktitle={Proceedings of the 23rd ACM SIGKDD international conference on knowledge discovery and data mining},
  pages={1027--1035},
  year={2017}
}

@inproceedings{blake2014marketplace,
  title={Why marketplace experimentation is harder than it seems: The role of test-control interference},
  author={Blake, Thomas and Coey, Dominic},
  booktitle={Proceedings of the fifteenth ACM conference on Economics and computation},
  pages={567--582},
  year={2014}
}

@article{fradkin2019simulation,
  title={A simulation approach to designing digital matching platforms},
  author={Fradkin, Andrey},
  journal={Boston University Questrom School of Business Research Paper Forthcoming},
  year={2019}
}

@article{holtz2020reducing,
  title={Reducing interference bias in online marketplace pricing experiments},
  author={Holtz, David and Lobel, Ruben and Liskovich, Inessa and Aral, Sinan},
  journal={Available at SSRN 3583836},
  year={2020}
}

@article{hernan2020causal,
  title={Causal Inference: What if. Boca Raton: Chapman \& Hill/CRC},
  author={Hernan, MA and Robins, J},
  year={2020}
}

@article{johari2022experimental,
  title={Experimental design in two-sided platforms: An analysis of bias},
  author={Johari, Ramesh and Li, Hannah and Liskovich, Inessa and Weintraub, Gabriel Y},
  journal={Management Science},
  year={2022},
  publisher={INFORMS}
}

@article{li2021interference,
  title={Interference, bias, and variance in two-sided marketplace experimentation: Guidance for platforms},
  author={Li, Hannah and Zhao, Geng and Johari, Ramesh and Weintraub, Gabriel Y},
  journal={arXiv preprint arXiv:2104.12222},
  year={2021}
}

@article{zigler2018bipartite,
  title={Bipartite Causal Inference with Interference},
  author={Zigler, Corwin M and Papadogeorgou, Georgia},
  journal={arXiv preprint arXiv:1807.08660},
  year={2018}
}

@article{harshaw2023design,
  title={Design and analysis of bipartite experiments under a linear exposure-response model},
  author={Harshaw, Christopher and S{\"a}vje, Fredrik and Eisenstat, David and Mirrokni, Vahab and Pouget-Abadie, Jean},
  journal={Electronic Journal of Statistics},
  volume={17},
  number={1},
  pages={464--518},
  year={2023},
  publisher={The Institute of Mathematical Statistics and the Bernoulli Society}
}

@article{shi2024scalable,
  title={Scalable Analysis of Bipartite Experiments},
  author={Shi, Liang and Bakhitov, Edvard and Hung, Kenneth and Karrer, Brian and Walker, Charlie and Bhole, Monica and Schrijvers, Okke},
  journal={arXiv preprint arXiv:2402.11070},
  year={2024}
}

@inproceedings{deng2013improving,
  title={Improving the sensitivity of online controlled experiments by utilizing pre-experiment data},
  author={Deng, Alex and Xu, Ya and Kohavi, Ron and Walker, Toby},
  booktitle={Proceedings of the sixth ACM international conference on Web search and data mining},
  pages={123--132},
  year={2013}
}

@article{stine1989introduction,
  title={An introduction to bootstrap methods: Examples and ideas},
  author={Stine, Robert},
  journal={Sociological Methods \& Research},
  volume={18},
  number={2-3},
  pages={243--291},
  year={1989},
  publisher={Sage Publications}
}

@article{tripuraneni2021meta,
  title={Meta-analysis of randomized experiments with applications to heavy-tailed response data},
  author={Tripuraneni, Nilesh and Madeka, Dhruv and Foster, Dean and Perrault-Joncas, Dominique and Jordan, Michael I},
  journal={arXiv preprint arXiv:2112.07602},
  year={2021}
}


\end{document}